\documentclass{article} 
\usepackage{collas2026_conference,times}
\usepackage{easyReview}

\usepackage{hyperref}
\hypersetup{
    colorlinks=true,
    linkcolor=red,
    filecolor=magenta,
    urlcolor=blue,
    citecolor=purple,
    pdftitle={Overleaf Example},
    pdfpagemode=FullScreen,
    }

\usepackage{amsmath,amssymb,amsfonts}
\usepackage{algpseudocode,algorithm}
\usepackage{subcaption}
\usepackage{graphicx}
\usepackage{wrapfig2}

\title{Flexible Empowerment at Reasoning \\with Extended Best-of-N Sampling}

\author{Taisuke Kobayashi \thanks{\url{https://prinlab.org/en/}} \\
National Institute of Informatics and\\
The Graduate University for Advanced Studies, SOKENDAI\\
Japan \\
\texttt{kobayashi@nii.ac.jp} \\
}

\preprintcopy 

\begin{document}

\maketitle

\begin{abstract}
This paper proposes a novel method that incorporates empowerment when reasoning actions in reinforcement learning (RL), thereby achieving the flexibility of exploration-exploitation dilemma (EED).
In previous methods, empowerment for promoting exploration has been provided as a bonus term to the task-specific reward function as an intrinsically-motivated RL.
However, this approach introduces a delay until the policy that accounts for empowerment is learned, making it difficult to adjust the emphasis on exploration as needed.
On the other hand, a trick devised for fine-tuning recent foundation models at reasoning, so-called best-of-N (BoN) sampling, allows for the implicit acquisition of modified policies without explicitly learning them.
It is expected that applying this trick to exploration-promoting terms, such as empowerment, will enable more flexible adjustment of EED.
Therefore, this paper investigates BoN sampling for empowerment.
Furthermore, to adjust the degree of policy modification in a generalizable manner while maintaining computational cost, this paper proposes a novel BoN sampling method extended by Tsalis statistics.
Through toy problems, the proposed method's cability to balance EED is verified.
In addition, it is demonstrated that the proposed method improves RL performance to solve complex locomotion tasks.
\end{abstract}

\section{Introduction}

For lifelong agents in the real world, such as autonomout robots, (continual) reinforcement learning (RL) \citep{sutton2018reinforcement,kumar2025continual} is one of the promissing methodologies due to its autonomy.
With some explicit goals defined as a scalar reward function, an agent can be expected to interact with an unknown environment to collect numerous experiences, and learn its optimal policy for achieving the goals.
That said, the learning efficiency is by no means high, and all practical RL applications so far have relied on simulations to artificially improve data collection efficiency \citep{silver2017mastering,rudin2022learning,hu2023simulation,radosavovic2024real}.
This approach is considered unsuitable for the goal of continual learning in the real world with the limited capability of simulations.

One reason for the poor learning efficiency in RL is that exploration is stochastic.
Essentially, RL agents act by randomly selecting the action, which might be around optimal to the best of current experiences; therefore, discovering a better solution requires relying on chance (unless the rewards and states are carefully designed to guide the agent toward a global optimum).
To overcome this inefficiency, additional strategies are needed that aim to explore and make new discoveries \citep{amin2021survey}, rather than simply maximizing the given reward.
This paper therefore focuses on one such strategy, \textit{empowerment}, which prioritizes the exploration of uncertain states where the agent’s actions are likely to make a significant contribution for the transitions to such states \citep{klyubin2005all}.

However, when incorporating such exploration strategies (including stochastic exploration) into RL, one must consider \textit{exploration-exploitation dilemma} (EED) \citep{thrun1992efficient}.
In other words, simply interacting with the environment by selecting actions predicted as the current best rarely lead to the discovery of better solutions.
On the other hand, focusing solely on exploration would prevent the goals from being fully achieved.
Since the number of action opportunities for the same situations is limited especially in lifelong agents\citep{kumar2025continual}, these two factors are inevitably in a trade-off, and an appropriate balance must be required to improve RL performance.

Even with conventional methods, this balance poses a problem when implementing RL algorithms in practice.
As one exploration strategy, the random selection of actions from a (simple) policy model has been replaced with a process that derives actions through (complex) transformations applied to randomly selected random variables \citep{plappert2018parameter,eberhard2023pink,chiappa2023latent,wang2023diffusion}.
In these approaches, annealing of randomness is a well-known method for adjusting EED, while it has been shown that this adjustment is theoretically difficult \citep{cesa2017boltzmann}.
It is particularly unsuitable for lifelong agents, which can operate in non-stationary environments.

A large number of strategies that generate active exploration are classified as intrinsically-motivated RL \citep{chentanez2004intrinsically}, incorporating a bonus term into the original reward to promote exploration.
To date, various metrics that define bonus terms (including empowerment) have been proposed \citep{bellemare2016unifying,burda2019exploration,pathak2019self,bharadhwaj2022information,yuan2022renyi}, and their usefulness is widely recognized.
However, even if the weight of this term is adjusted according to the learning context \citep{haarnoja2018soft2,kobayashi2023reward,hugessen2024surprise}, a delay occurs because the value function must first be learned before the adjustment takes effect.
Furthermore, for bonus terms that do not theoretically maintain the form of reward shaping \citep{ng1999policy}, the policy may converge to one that differs from the one intended to achieve the original objective.
If this intrinsically-motivated RL is extended to a multi-objective RL framework \citep{morere2018bayesian,al2022getting}, the weights of such exploration bonuses could be adjusted even after the value function has been learned, allowing for the derivation of policies that achieve the objectives.
However, because the bonuses are non-stationary in most cases, estimating their value functions are difficult to be estimated, leaving significant concerns regarding accuracy.

Given the above background, this paper focuses on the best-of-N (BoN) sampling trick, a policy adjustment used at reasoning that has garnered significant attention in recent years \citep{stiennon2020learning}, as a method for incorporating different exploration strategies.
This trick involves randomly selecting $N$ random variables from a source distribution and choosing the one that maximizes the objective function (not limited to the reward function in general).
The final selected random variable can be regarded as having been sampled from a distribution that differs from the source one and is biased by the objective function.
By utilizing this trick, since the source distribution, the policy in RL, prioritizes exploitation, setting a metric that promotes exploration (empowerment in this paper) as the objective of BoN sampling allows us to implicitly obtain another policy that prioritizes exploration.

However, the degree to which BoN sampling favors exploration depends on the sample size $N$, and computational costs fluctuate accordingly.
For agents such as robots operating in the real world, this fluctuation can cause problems such as controller instability and response delays.
In this paper, to avoid this issue, BoN sampling is theoretically extended, so-called E-BoN sampling.
This method builds upon the softmax strategy to BoN sampling \citep{verdun2025soft,ichihara2025evaluation} and introduces the entmax strategy \citep{peters2019sparse,gonccalves2025adasplash} by utilizing Tsalis statistics \citep{tsallis1988possible}.
As a result, EED balance can be adjusted using a new parameter, while simultaneously enabling a robust implementation that accounts for the state-dependence of empowerment.
Furthermore, instead of iterative solvers needed for the conventional entmax implementation, a new accurate approximation is developed for making computational costs constant.

Through numerical experiments conducted with dm\_control \citep{tunyasuvunakool2020dm_control} as benchmark, it is demonstrated that the proposed E-BoN sampling with empowerment possesses the flexibility of EED.
Furthermore, by introducing a simple sample-based strategy for leveraging this flexibility, it achieves well-balanced locomotion control across tasks compared to other baselines.
In summary, the contributions of this papers are thre folds:
\begin{enumerate}
    \item E-BoN sampling, a theoretical extension of BoN sampling, was derived for the first time in the world;
    \item its practical implementation was also designed with distinct roles for the respective parameters and the approximated solution with constant computation cost;
    \item and the flexibility of EED and benefits to RL performance of the proposed E-BoN sampling with empowerment were experimentally demonstrated.
\end{enumerate}

\section{Preliminaries}

\subsection{Markov decision process}

This paper deals with Markov decision process (MDP), a basic RL problem setting \citep{sutton2018reinforcement}.
While it is possible to apply the proposed method to partially observable MDP and other variants, they are omitted here to simplify the disussion and implementation.

MDP is defined with the following tuple: $(\mathcal{S}, \mathcal{A}, \mathcal{R}, p_e, p_0)$.
$\mathcal{S} \subseteq \mathbb{R}^{|\mathcal{S}|}$ is the state space, $\mathcal{A} \subseteq \mathbb{R}^{|\mathcal{A}|}$ is the action space, $\mathcal{R} \subseteq \mathbb{R}$ is the reward set, $p_e$ and $p_0$ are the state-transition and initial-state probabilities of a target environment, respectively.
At time step $t$, the state $s_t \in \mathcal{S}$ is sampled from $p_e$ (or $p_0$ if $t=0$).
An agent then selects an action $a_t \in \mathcal{A}$ from its trainable policy $\pi$ according to $s_t$.
By the interaction with $a_t$, the environment probabilistically transitions to the next state $s_{t+1}$ (or $s_t^\prime$ for ease of distinction) according to $p_e$.
This single transition is evaluated as $r_t = r(s_t, a_t, s_{t+1}) \in \mathcal{R}$, which represents the task quality.

The cumulative reward over time by repeating transitions is called the return, and is typically defined using a discount factor $\gamma \in [0, 1)$ as $R_t = \sum_{k=0}^\infty \gamma^k r_{t+k}$.
In RL, the goal is to optimize $\pi$ (more specifically, its parameters $\phi$) such that the expected $R_t$ is maximized at every time step.
Note that, since information about $t$ is often unnecessary in the following discussion, it is often omitted in this paper.

\subsection{Soft actor-critic algorithm}
\label{subsec:sac}

As will be discussed later, the proposed method is designed for off-policy RL algorithms.
In this paper, soft actor-critic (SAC), which is one of the most popular algorithms in decade \citep{haarnoja2018soft}, is introduced as a baseline.
Although SAC itself facilitates exploration through the maximization of policy entropy, if automatic temperature tuning \citep{haarnoja2018soft2} is applied, its primary purpose is to solve the given task; therefore, the exploration promotion provided by the proposed method remains effective.
Furthermore, since a policy with maximized entropy has the potential to generate diverse actions, it is effective as a source distribution for BoN sampling (described later).
Given the suitability of SAC, while the combination of the proposed method with other off-policy RL algorithms should be investigated in the future, this paper will evaluate the proposed method exclusively using SAC.

In SAC, the entropy term of the policy, $\mathcal{H}(\pi)$, to the reward $r$ associated with the task provided by the environment.
\begin{align}
    r \gets r + \alpha^\mathrm{SAC} \mathcal{H}(\pi(\cdot \mid s; \phi))
\end{align}
where, $\alpha^\mathrm{SAC} \geq 0$ denotes the temperature parameter, which is automatically tuned even in this paper.
The action-value function $Q(s,a; \theta) = \mathbb{E}[\sum_{k=0}^\infty r_{t+k} + \alpha^\mathrm{SAC} \mathcal{H}(\pi(\cdot \mid s_{t+k}; \phi)) \mid s_t=s, a_t=a]$ ($\theta$ denotes the trainable parameters) follows the soft Bellman equation, deriving the following loss function to be minimized.
\begin{align}
    \mathcal{L}(\theta) = \mathbb{E}_{(s,a,s^\prime,r) \sim D}\left[ \left\{ r + \gamma \mathrm{sg}(Q(s^\prime, a^\prime; \theta) - \alpha^\mathrm{SAC} \ln \pi (a^\prime \mid s^\prime; \phi)) - Q(s, a; \theta) \right\}^2 \right]
\end{align}
where, $D$ denotes a replay buffer, $\mathrm{sg}(\cdot)$ means the stop gradient, and $a^\prime$ is sampled from $\pi(\cdot \mid s^\prime; \phi)$.
Since the theoretically optimal policy is known as $\pi^\ast(a \mid s) = \exp(Q(s,a)/\alpha^\mathrm{SAC}) / Z(s)$ (where $Z(s)$ is the (implicit) normalization constant), the policy model $\pi(a \mid s; \phi)$, which can easily generate actions, is learned by minimizing the following Kullback-Leibler (KL) divergence.
\begin{align}
    \mathcal{L}(\phi) = \mathbb{E}_{s \sim D}\left[ \mathrm{KL}(\pi(\cdot \mid s; \phi) \mid \pi^\ast(\cdot \mid s)) \right]
    \propto \mathbb{E}_{s \sim D}\left[ - Q(s, a; \theta) + \alpha^\mathrm{SAC} \ln \pi (a \mid s; \phi) \right]
\end{align}
where, $a$ is sampled from $\pi(\cdot \mid s; \phi)$, not from $D$.
In practice, for $Q$, double Q-learning with target networks are usually introduced to stabilize learning (see Appendix~\ref{app:config}).

\subsection{Empowerment}

This paper introduces empowerment \citep{klyubin2005all} as a strategy that promotes exploration under MDP.
Specifically, let us first consider the following conditional mutual information.
\begin{align}
    \mathcal{I}(s_{t+H}, A_t^H \mid s_t) &= \mathbb{E}_{p(s_{t+H}, A_t^H \mid s_t)} \left[ \ln \frac{p(s_{t+H}, A_t^H \mid s_t)}{p(A_t^H \mid s_t) p(s_{t+H} \mid s_t)} \right]
    \nonumber \\
    &= \mathbb{E}_{p(s_{t+H}, A_t^H \mid s_t)} \left[ \ln p(s_{t+H}, A_t^H \mid s_t) - \ln p(A_t^H \mid s_t) - \ln p(s_{t+H} \mid s_t) \right]
    \label{eq:def_mi}
\end{align}
where, $H \geq 1$ denotes the prediction horizon, and $A_t^H = \{a_{t+k}\}_{k=0}^{H-1}$ is the sequence of actions.
This mutual information shows the extent to which the sequence of actions contributed to the state transition from $s_t$ to $s_{t+H}$.
If MDP holds, that is, if $s_{t+H}$ is determined stochastically by a chain of $\pi$ and $p_e$, the above $p(s_{t+H}, A_t^H \mid s_t)$ and $p(A_t^H \mid s_t)$ can be decomposed using $p(S_t^H \mid s_t, A_t^H) = \prod_{k=0}^{H-2} p_e(s_{t+k+1} \mid s_{t+k}, a_{t+k})$ with $S_t^H = \{s_{t+k}\}_{k=1}^{H-1}$ as follows:
\begin{align}
    \begin{split}
        p(s_{t+H}, A_t^H \mid s_t) &= \mathbb{E}_{p(S_t^H \mid s_t, A_t^H)}\left[ \left\{ \prod_{k=0}^{H-1} \pi(a_{t+k} \mid s_{t+k}; \phi) \right\} p(S_t^H \mid s_t, A_t^H) p_e(s_{t+H} \mid s_{t+H-1}, a_{t+H-1}) \right]
        \\
        p(A_t^H \mid s_t) &= \mathbb{E}_{p(S_t^H \mid s_t, A_t^H)}\left[ \left\{ \prod_{k=0}^{H-1} \pi(a_{t+k} \mid s_{t+k}; \phi) \right\} p(S_t^H \mid s_t, A_t^H) \right]
    \end{split}
\end{align}
Then, $\mathcal{I}(s_{t+H}, A_t^H \mid s_t)$ can be approximated and simplified as follows:
\begin{align}
    \eqref{eq:def_mi} &\simeq \mathbb{E}_{p(s_{t+H}, A_t^H \mid s_t)} \left[ \mathbb{E}_{p(S_t^H \mid s_t, A_t^H)}\left[ \ln p_e(s_{t+H} \mid s_{t+H-1}, a_{t+H-1}; \phi) \right] - \ln p(s_{t+H} \mid s_t) \right]
    \nonumber \\
    &= \mathbb{E}_{p(s_{t+H}, A_t^H \mid s_t) p(S_t^H \mid s_t, A_t^H)} \left[ \ln p_e(s_{t+H} \mid s_{t+H-1}, a_{t+H-1}; \phi) - \ln p(s_{t+H} \mid s_t) \right]
    \nonumber \\
    &= \mathbb{E}_{p(\tau_t \mid s_t; \phi)} \left[ \mathrm{KL}\left( p_e(s_{t+H} \mid s_{t+H-1}, a_{t+H-1}; \phi) \mid p(s_{t+H} \mid s_t) \right) \right]
\end{align}
where, $p(\tau_t \mid s_t; \phi) = \prod_{k=0}^{H-1} \pi(a_{t+k} \mid s_{t+k}; \phi) p(S_t^H \mid s_t, A_t^H)$ with $\tau_t$ the state-action trajectory.
The reason the additional condition $\phi$ is added to $p_e$ is to explicitly represent its indirect influence on state transitions.
Note that with $H=1$, this simplification holds exactly rather than approximately, and the same is true if this mutual information is modified to the discounted one~\citep{schneider2025information}.

Empowerment refers to the maximum mutual information obtained through policy optimization.
\begin{align}
    \mathcal{E}(s_t) = \max_{\phi} \mathbb{E}_{p(\tau_t \mid s_t)} \left[ \mathrm{KL}\left( p_e(s_{t+H} \mid s_{t+H-1}, a_{t+H-1}; \phi) \mid p_m(s_{t+H} \mid s_t) \right) \right]
    \label{eq:def_emp}
\end{align}
Note that the agent has no state-transition model originally, $p_e$ and $p_m$ (as a marginal model) need to be modeled with parameters $\psi$, which can be optimized by supervised learning (see Appendix~\ref{app:model}).
With this definition, the empowered policy can be basically interpreted as aiming for a state where, despite higher uncertainty, the agent’s own generated actions contribute significantly to the transition, thereby reducing uncertainty.
Alternatively, due to the effects of modeling errors, the agent may prioritize states where its intervention increases uncertainty (in order to learn correctly).
In any case, by incorporating this form of empowerment as an objective, the agent's exploration can be facilitated.

\subsection{Best-of-N sampling}

In this study, BoN sampling \citep{stiennon2020learning} is the main target for obtaining the desired policy during reasoning time without training.
Specifically, BoN requires the objective function $J(x)$ to be maximized, where $x$ is a random variable sampled from its sampler (also known as the source distribution), $p_x$.
Then, $N$ candidates of $x$ are randomly generated from $p_x$.
\begin{align}
    x_i \sim p_x, \ i=1,\ldots,N
\end{align}
From them, the desired one is greedy selected.
\begin{align}
    x^\ast = \arg\max_{x_i} J(x_i)
\end{align}
Intuitively, if $N$ is sufficiently large, the true maximum objective can be found; otherwise, the sampling bias is remained.
In other words, by adjusting $N$, the degree of divergence between the source and tilted distributions\footnote{In this paper, the degree of divergence corresponds to the balance between exploration via empowerment maximization and exploitation via a base policy that maximizes the value function.} can be balanced.
However, the computational cost also varies proportionally with $N$, even if parallel computation on GPU is adopted.
Since this directly affects latency, BoN sampling with adaptive $N$ is unsuitable for problems that require the stable latency, such as real-time robot control.

Soft BoN (S-BoN in this paper) sampling \citep{verdun2025soft,ichihara2025evaluation} has been proposed as a method for adjusting the deviation in a different way.
As the name suggests, whereas the basic BoN sampling involved a hard selection using $\arg\max$ (therefore, it is named H-BoN in this paper), this method performs the selection probabilistically using a softmax strategy.
Specifically, the tilted distribution that generates $x^\ast$ by BoN sampling, $p_x^\ast$, is regarded as the solution of the following constrained optimization problem.
\begin{align}
    p_x^\ast = \arg\max_{p_x^\prime} \mathbb{E}_{p_x^\prime}[J(x)], \ \mathrm{s.t.} \ \mathrm{KL}(p_x^\prime \mid p_x) \leq \epsilon
    \label{eq:prb_sbon}
\end{align}
where, $\epsilon \geq 0$ denotes the threshold, which is known to increase with $N$.
This optimization problem can be reformulated by combining Lagrange's method of undetermined multipliers (with $1/\beta$ as the Lagrange multiplier) with importance sampling trick and Monte Carlo approximation, as follows:
\begin{align}
    p_x^\ast &= \arg\max_{p_x^\prime} \mathbb{E}_{p_x}\left[ \frac{p_x^\prime}{p_x} \beta J(x) - \frac{p_x^\prime}{p_x} \ln \frac{p_x^\prime}{p_x} \right]
    \nonumber \\
    p_x^\ast &\simeq \arg\max_{p_x^\prime} \sum_{i=1}^N P_i^\prime \beta J(x_i) + \mathcal{H}(P_{1:N}^\prime)
    \label{eq:prb_softmax}
\end{align}
where, $P_i^\prime = p_x^\prime(x_i) / (N p_x(x_i))$ with $\sum_{i=1}^N P_i^\prime = 1$.

This optimization problem is equivalent to finding the optimal probability $P_{1:N}$ and selecting $x^\ast \in \{x_i\}_{i=1}^N$ in accordance with it.
Note that, due to the presence of the entropy term, $P_{1:N}$ tends toward a uniform probability, leading to $p_x^\prime = p_x$.
Anyway, it is known that $P_{1:N}$ is given by the softmax function.
\begin{align}
    P_i = \frac{e^{\beta J(x_i)}}{\sum_{j=1}^N e^{\beta J(x_j)}}
\end{align}
where, $\beta \geq 0$ is now regarded as the inverse temperature parameter.
Using this, $x^\ast$ is selected probabilistically as follows:
\begin{align}
    x^\ast \sim \mathcal{C}(P_{1:N})
\end{align}
where, $\mathcal{C}$ denotes the categorical distribution (to be precise, the index $i$ is sampled from it and $x_i$ is denoted as $x^\ast$).

With this S-BoN sampling, reducing $\beta$ increases the randomness, resulting in $p_x^\ast \simeq p_x$; conversely, increasing $\beta$ causes it to asymptotically approach the one obtained by H-BoN sampling.
In other words, even if $N$ is fixed, as long as $N$ is sufficiently large within the acceptable latency range, $p_x^\ast$ can be adjusted during reasoning with a constant latency.
However, as can be seen from the fact that $\beta$ can be interpreted as a scaling factor of $J(x)$, the appropriate $\beta$ depends heavily on the design of $J(x)$.
If $J(x)$ is defined using a state-dependent metric such as empowerment, its scale can easily vary depending on the state the agent is facing, making it difficult to adjust $\beta$ by hand.

\section{Proposal}

\subsection{Overview}

\begin{minipage}{0.48\linewidth}
    \begin{algorithm}[H]
        \caption{Basic pseudocode}
        \label{alg:baseline}
        \begin{algorithmic}[1]
            \State{Initialize parameters $\theta$, $\phi$ and replay buffer $D$}
            \State{Set sampling strategy $\Pi$ with parameters $\psi$}
            \While{not converged}
                \While{not terminated or truncated}
                    \State{Get the current state $s$}
                    \State{Decide the action with $\Pi(s, \pi(\cdot \mid s; \phi); \psi)$}
                    \State{Get the next state $s^\prime$ and reward $r$}
                    \State{Store experience $D = D \cup (s, a, s^\prime, r)$}
                \EndWhile
                \State{Update $\theta$, $\phi$ by off-policy RL algorithm with $D$}
                \State{Update $\psi$ with $D$ if necessary}
            \EndWhile
        \end{algorithmic}
    \end{algorithm}
\end{minipage}
\begin{minipage}{0.48\linewidth}
    \begin{algorithm}[H]
        \caption{E-BoN sampling strategy}
        \label{alg:proposal}
        \begin{algorithmic}[1]
            \State{\textbf{Parameter:} \#samples $N$ and EED balance $\alpha$}
            \State{\textbf{Input:} $\pi(\cdot \mid s; \phi)$, $s$, $\psi$}
            \State{Set the objective function $J(a, s; \psi): \mathcal{A} \times \mathcal{S} \mapsto \mathbb{R}_+$}
            \State{Sample $\{a_i\}_{i=1}^N \sim \pi(a \mid s; \phi)$}
            \State{Compute $J_i = J(a_i, s; \psi)$ for $i=1,\ldots,N$}
            \State{Compute $\beta = \cfrac{1}{N^{-1}\sum_{j=1}^N J_j}$}
            \State{Scale $J_i \gets \beta J_i$}
            \State{Solve entmax of $J_i$ to get $P_i$ for $i=i,\ldots,N$}
            \State{Sample $a \sim \mathcal{C}(P_{1:N})$}
            \State{\Return $a$}
        \end{algorithmic}
    \end{algorithm}
\end{minipage}

The basic pseudocode for this study is shown in Alg.~\ref{alg:baseline}.
The key point is the method for determining actions when interacting with the environment (Line~6).
While based on the policy being trained, $\pi$, actions are determined by implicitly modifying it to a different policy, $\tilde{\pi}$, in accordance with empowerment.
In other words, since $\tilde{\pi} \neq \pi$, this modification is permitted only in off-policy RL algorithms.
Note that when adding an exploration-promoting term to the task-specific reward, as in conventional methods, the total reward function becomes non-stationary, making this approach suitable for on-policy RL algorithms (although it actually functions without issue in off-policy RL algorithms as well).

In the proposed method, Alg.~\ref{alg:proposal} is applied to the target Line~6.
As mentioned above, in the basic H-BoN sampling, balancing EED requires changing the number of samples $N$, making it impossible to keep the computational cost constant.
Although the improved S-BoN sampling allows for EED balancing with constant computational cost via the inverse temperature parameter $\beta$, in practice, it is difficult to configure settings suitable for empowerment, where the scale can easily fluctuate depending on the faced state.
To address these issues, this paper further extends S-BoN sampling and proposes E-BoN sampling, which utilizes the entmax strategy \citep{peters2019sparse}.
E-BoN sampling introduces a new scalar $\alpha \in \mathbb{R}$ that determines the deformation of the exponential and logarithmic functions defined in Tsalis statistics \citep{tsallis1988possible}.
EED can be balanced even by $\alpha$.
Although this may appear to duplicate the functionality of $\beta$, the addition of $\alpha$ allows $\beta$ to be used for adaptively scaling empowerment.
That is, $\alpha$ can focus exclusively on balancing EED, although a straightforward implementation is difficult.
These differences are summarized in Table ~\ref{tab:comparison}, and a detailed implementation of E-BoN sampling is described below.

\begin{table}[ht]
    \caption{Comparison of sampling strategies for balancing EED}
    \label{tab:comparison}
    \centering
    {\scriptsize
    \begin{tabular}{lcccc}
        Strategy & $N$ & $\beta$ & $\alpha$ & Drawback
        \\ \hline \\
        Random sampling & N/A & N/A & N/A & No capability to balance EED
        \\
        H-BoN sampling \citep{stiennon2020learning} & Adjusted & N/A & N/A & Non-constant computational cost
        \\
        S-BoN sampling \citep{verdun2025soft,ichihara2025evaluation} & Fixed & Adjusted & N/A & Sensitivity to scale of empowerment
        \\
        E-BoN sampling (Proposal) & Fixed & Adapted for scaling & Adjusted & Complex implementation
        \\
    \end{tabular}
    }
\end{table}

\subsection{From softmax to entmax: extension of BoN sampling}

First, by introducing Tsalis statistics \citep{tsallis1988possible}, extension of the standard Gibbs-Shannon statistics, into the derivation of S-BoN sampling the basis of E-BoN sampling is developed.
As a preliminary step, the following q-exponential and q-logarithmic functions are introduced.
\begin{align}
    \exp_q(x) = [1 + (1-q)x]_+^{\frac{1}{1-q}}, \ \ln_q(x) = \frac{x^{1-q} - 1}{1-q}
\end{align}
Although $q \in \mathbb{R} \\ 1$ is the proper domain, $\exp_q \to \exp$ and $\ln_q \to \ln$ hold when $q \to 1$.
Hereinafter, for the sake of simplicity, $\exp_1 = \exp$ and $\ln_1 = \ln$ are assumed, making $q$ belong to $\mathbb{R}$.

Using these, the constraint in eq.~\eqref{eq:prb_sbon}, which serves as the starting point of S-BoN sampling, is replaced as follows:
\begin{align}
    \mathrm{KL}(p_x^\prime \mid p_x) \leq \epsilon \to \mathrm{KL}_{\alpha+1}(p_x^\prime \mid p_x) \leq \epsilon_\alpha
\end{align}
where, $\alpha \in \mathbb{R}$ denotes the hyperparameter.\footnote{This definition is biased from the paper introducing entmax \citep{peters2019sparse} for prioritizing ease of use.}
In addition, $\mathrm{KL}_{q}(p_1 \mid p_2) = \mathbb{E}_{x\sim p_1}\left[ - \ln_q \frac{p_2(x)}{p_1(x)} \right]$ is q-deformed KL (or Tsallis) divergence.
Note that since the threshold assigned implicitly should vary depending on $\alpha$, it is explicitly distinguished here by denoting it as $\epsilon_\alpha$.

Following the same procedure as in the case of S-BoN sampling, this optimization problem is reformulated as selecting $x^\ast$ from $N$ candidates $\{x_i\}_{i=1}^N$, which are sampled by $p_x$.
\begin{align}
    p_x^\ast &= \arg\max_{p_x^\prime} \mathbb{E}_{p_x}\left[ \frac{p_x^\prime}{p_x} \beta J(x) + \frac{p_x^\prime}{p_x} \ln_{\alpha+1} \frac{p_x}{p_x^\prime} \right]
    \nonumber \\
    p_x^\ast &\simeq \arg\max_{p_x^\prime} \sum_{i=1}^N P_i^\prime \beta J(x_i) + \mathcal{H}_{\alpha+1}(P_{1:N}^\prime)
    \label{eq:prb_entmax}
\end{align}
where, $\mathcal{H}_q(p) = \mathbb{E}_{x \sim p}\left[ \ln_q \frac{1}{p(x)} \right]$ is Tsallis entropy.\footnote{The definition in the paper introducing entmax \citep{peters2019sparse} was wrong probably due to simplify the derivation.}
Under the constraint about $\sum_{i=1}^N P_i = 1$, this can be solved as below.
\begin{align}
    P_i = \exp_{1-\alpha}(\beta (\alpha + 1)^{-1} J(x_i) - \lambda) = [1 + \alpha (\beta J(x_i) - \lambda)]_+^{\frac{1}{\alpha}}
    \label{eq:res_entmax}
\end{align}
where, $\lambda$ is the Lagrange multiplier for the constraint of summation, and $(\alpha + 1)^{-1}$ is cancelled out by $\beta$ (otherwise, $\alpha > -1$ is required for sampling the better one).
That is, by solving $\lambda$, $\sum_{i=1}^N P_i = 1$ holds, while it cannot be analytically solved.
Originally, $\lambda$ has been found using iterative numerical solvers; however, since keeping the computational cost constant is crucial in this paper, an alternative approximate method is developed (see Appendix~\ref{app:entmax}).

\begin{figure}[ht]
    \centering
    \includegraphics[keepaspectratio=true,width=0.96\linewidth]{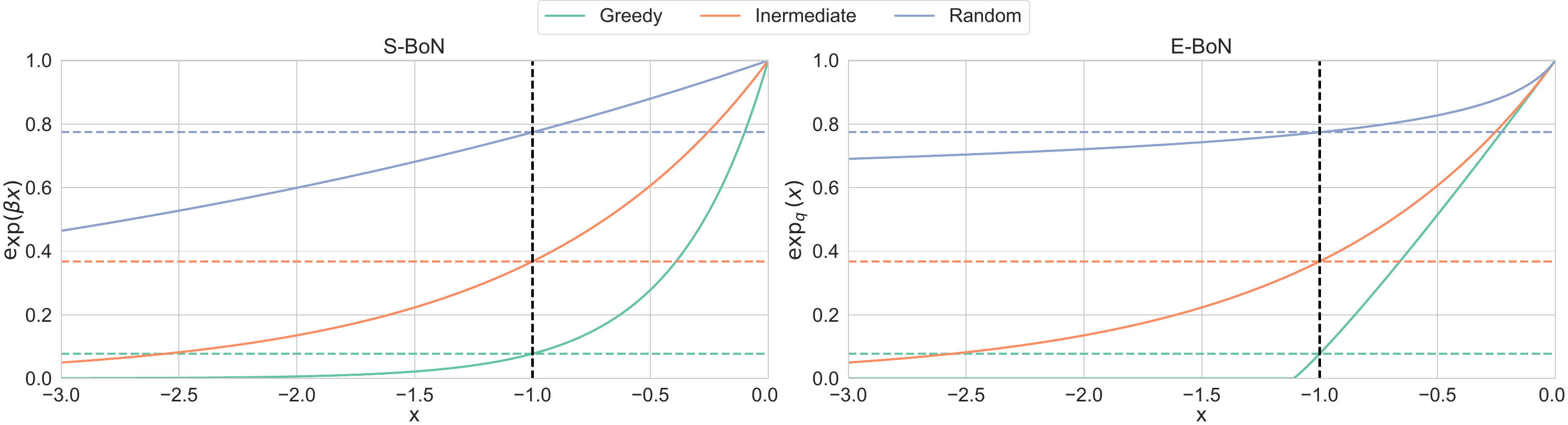}
    \caption{Comparison of probability shpaes in S-BoN sampling and E-BoN sampling}
    \label{fig:effect_beta_alpha}
\end{figure}

The differences between this E-BoN sampling and the conventional S-BoN sampling are now visualized in Fig.~\ref{fig:effect_beta_alpha}.
Note that the maximum value of $x$ in the figure is set to zero because $\lambda \geq \max_i x_i$, and because the softmax function in S-BoN sampling also subtracts $\max_i x_i$ to stabilize the numerical computation.
The left chart shows the exponential function used in S-BoN sampling adjusted by $\beta$, while the right chart shows the q-exponential function appearing in E-BoN sampling adjusted by $\alpha$ while fixing $\beta=1$ so that the value at $x=-1$ matches that of the left chart.
Naturally, when $\beta=1$, the two figures match with $\alpha=0$.
When $\beta > 1$ is used to accentuate the scale difference in $J$ and achieve a greedy selection, the probability remains non-zero in theory regardless of how relatively small $J$ is.
However, when $\alpha > 0$ is used to achieve a similarly greedy selection, values below a certain threshold are clipped to zero, eliminating the selection probability.
This is the sparsity that has been the focus of previous research on entmax; conversely, it can be seen that when $\alpha < 0$, the probability difference narrows.
In particular, compared to cases where $\beta < 1$ is used to increase randomness, the tail becomes heavier, making it easier to equally select candidates even with smaller values.
In summary, it can be concluded that adjusting $\alpha$ in E-BoN sampling using entmax allows for a wider range of selection, from more random (i.e. maintaining the source distribution) to more deterministic (i.e. asymptotically approaching H-BoN sampling), compared to adjusting $\beta$ in S-BoN sampling.

As a remark, the analyzed behaviors of E-BoN sampling w.r.t. $\alpha$ can also be explained qualitatively by the constraint imposed by Tsallis divergence mentioned at the beginning.
Since $\ln_{q_1}(x) \leq \ln_{q_2}(x)$ holds for $q_1 > q_2$, $\mathrm{KL}_{\alpha+1}(\cdot \mid \cdot) \geq \mathrm{KL}(\cdot \mid \cdot)$ holds for $\alpha > 0$ as well.
Since both are non-negative, $\mathrm{KL}_{\alpha+1}(\cdot \mid \cdot)$ has a larger curvature.
Therefore, even if $p_x^\ast$ retains some deviation from $p_x$, it becomes easier to satisfy the constraint (under the assumption of $\epsilon_{\alpha>0} > \epsilon$), which can be interpreted as enabling a greedy selection closer to H-BoN sampling.
Conversely, when $\alpha < 0$, $\mathrm{KL}_{\alpha+1}(\cdot \mid \cdot)$ has a small curvature, and $p_x^\ast$ must be brought closer to $p_x$ in order to satisfy the constraint (assumiing $\epsilon_{\alpha<0} < \epsilon$).

\subsection{Auto-scaling with non-negative Monte carlo approximation}

Although the above E-BoN sampling is considered a useful alternative to S-BoN sampling, it inherits the latter’s sensitivity to the scale of the objective function $J$.
A clear example of this is the threshold at which $P_i$ becomes zero for $\alpha > 0$, which can be easily identified as $\beta J(x_i) - \lambda \leq \alpha^{-1}$.
Depending on the scale of $J$, this can lead to situations where no options are excluded at all, or conversely, where most options are excluded.

Since $\beta$ is remained as a hyperparameter, E-BoN sampling aims to automatically adjust the scale by designing it.
A naive approach would be to scale using the max-min sample values of $J$ or the sample standard deviation.
However, in the former case, max-min scaling, while the range of $\beta J$ becomes closed and it becomes easier to define the range of sparsification for $\alpha > 0$, it fails to fully utilize the heavy tail for $\alpha < 0$.
Conversely, in the latter case of standard deviation scaling, while the heavy tail for $\alpha < 0$ can be utilized due to open range, it is difficult to predict how many candidates will exceed the threshold for $\alpha > 0$ because $J$ may not necessarily follow Gaussian.

This paper introduces a scaling method based on the mean value of $J$, which, despite its simple implementation, is considered to exhibit reasonable behaviors.
Specifically, $\beta$ is given simply as follows:
\begin{align}
    \beta = \frac{1}{N} \sum_{i=1}^N J(x_i)
\end{align}
In this case, since the range of $\beta J$ is not restricted, the heavy tail should be effectively utilized as well as the standard deviation scaling.
Furthermore, regarding sparsification, the skewness of $J$ can be reflected appropriately.
That is, if the candidates are biased toward $\max_i J(x_i)$, only a few clearly poor candidates are excluded; conversely, if they are biased toward $\min_i J(x_i)$, it becomes possible to select only from a subset of excellent candidates.

However, it is important to note that this scaling is valid only for $J \in \mathbb{R}_+$.
Since the empowerment of eq.~\eqref{eq:def_emp}, which is the subject of this paper, is the expectation of KL divergence (and its maximization), it theoretically satisfies this condition due to the non-negativity of KL divergence.
However, in practice, since a closed-form expression for KL divergence cannot be obtained depending on the models of the two target distributions, an approximate calculation is necessary.
The most popular method is Monte Carlo approximation, which involves, in the extreme case, generating a single sample and using the log-likelihood ratio of the two target distributions for that sample.
This is helpful for approximating the expectation in empowerment as well.
However, as easily imagined, this approximation cannot guarantee non-negativity.
Therefore, in this paper, the equivalent transformation of KL divergence proposed by \citet{nielsen2020non}, which guarantees non-negativity during Monte Carlo approximation, is adopted.

Finally, $J$ is defined as follows:
\begin{align}
    J(a_i, s; \psi) &= \ln \frac{p_e(s_i^\prime \mid s, a_i; \psi)}{p_m(s_i^\prime \mid s; \psi)} + \frac{p_m(s_i^\prime \mid s; \psi)}{p_e(s_i^\prime \mid s, a_i; \psi)} - 1
    \\
    \mathrm{s.t.}\ s_i^\prime &= \mathbb{E}_{s^\prime \sim p_e(s^\prime \mid s, a_i; \psi)}[s^\prime]
    \nonumber
\end{align}
where, $s_i^\prime$ is fixed to the mean of $p_e$ to stabilize the behavior, avoiding sampling.
For simplicity and computational efficiency, this paper adopts $H=1$ for empowerment.

\section{Experiments}

\subsection{Setup}

This paper addresses the following two research questions:
\begin{enumerate}
    \item Does the proposed E-BoN sampling have the ability to adjust EED?
    \item Is that effective in improving task performance?
\end{enumerate}
To address these questions, two types of experiments are conducted below.
The training conditions, such as the model architecture used in the experiments, are described in Appendix~\ref{app:config}, and the benchmark tasks are selected from dm\_control \citep{tunyasuvunakool2020dm_control} wrapped by Shimmy \citep{jun2023shimmy}.
Note that the maximum time steps per episode is basically limited to 500.
Furthermore, evaluation is based on the return (i.e. higher is better) obtained when actions are determined greedily using the post-trained policies.

In the first experiment, the flexibility of EED is investigated with \textit{dm\_control/cartpole-balance\_sparse-v0} (CartPole) and \textit{dm\_control/point\_mass-easy-v0} (PointMass) as distinct toy problems.
Note that only CartPole has the maximum time steps of 1000 to expose the adverse effects of unnecessary exploration by thinning out the data density near the initial state.
Since the former has an optimal initial state and the goal is to maintain it, performance improvements can be expected by pursuing return maximization without promoting exploration.
Conversely, since the latter is a task of moving from a random initial state to the goal, promoting exploration should be important for efficiently discovering the goal.
For these tasks, a total of nine conditions are compared: seven conditions for the proposed E-BoN sampling with $\alpha=\{-2, -1, -0.5, 0, 0.5, 1, 2\}$; Random sampling from $\pi$, which corresponds to $\alpha=-\infty$; and H-BoN sampling corresponding to $\alpha=\infty$.
From them, whether performance changes occur depending on $\alpha$ is first confirmed.
Then, if so, whether the trend of changes is led by balancing EED is considered.

In the second experiment, E-BoN sampling with balanced EED is compared to baselines in terms of learning performance.
As more difficult tasks than the above toy problems, three locomotion tasks are selected:
\textit{dm\_control/cheetah-run-v0} (Cheetah), \textit{dm\_control/walker-run-v0} (Walker), and \textit{dm\_control/quadruped-walk-v0} (Quadruped).
While efficient exploration is required due to the more complex dynamics, the wide variety of possible behaviors necessitates narrowing down the behavior by pursuing the maximization of return.
For comparison, in addition to the aforementioned Random sampling and H-BoN sampling, S-BoN sampling (i.e. the proposed implementation fixed at $\alpha=0$) are prepared as baselines.
The proposed E-BoN sampling has $\alpha \in [-2, 2]$ as variable.
Although meta-optimization techniques \citep{akiba2019optuna} are considered effective for optimizing $\alpha$, this paper adopts a simple strategy of episodically sampling $\alpha$ from an arcsine distribution, which prioritizes distinctive boundaries.
Even with this approach alone, the replay buffer will contain a mix of data that prioritize either exploration or exploitation, from which efficient learning can be expected.

\subsection{Flexibility of EED}

\begin{figure}[ht]
    \centering
    \includegraphics[keepaspectratio=true,width=0.96\linewidth]{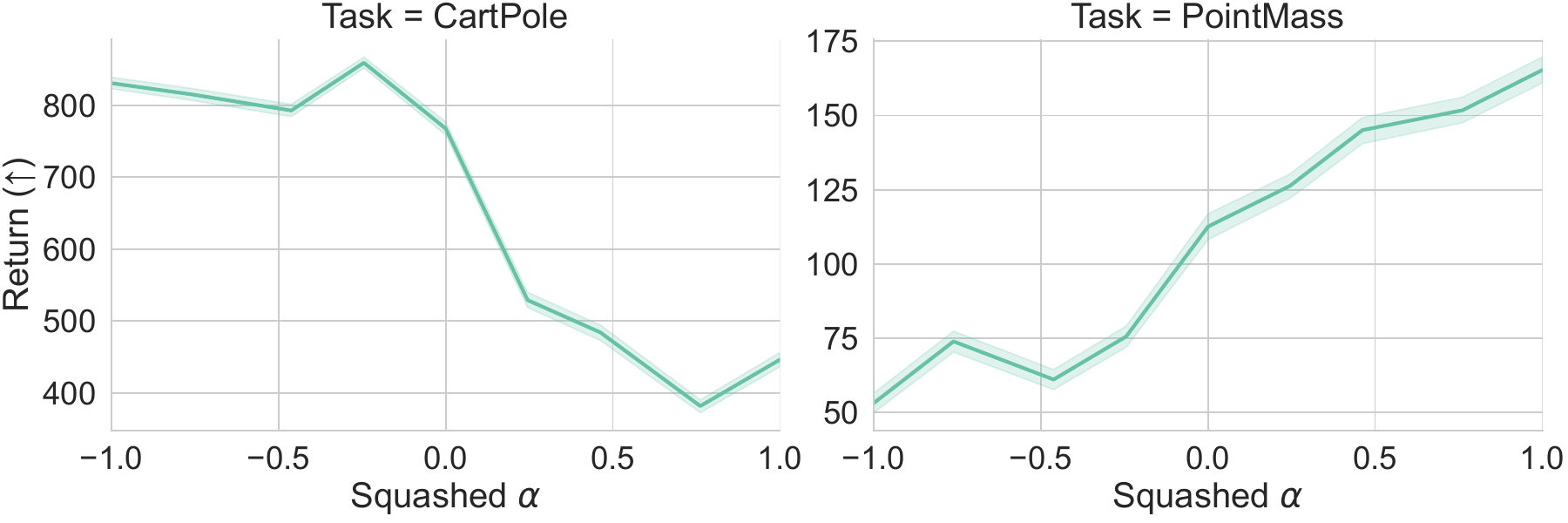}
    \caption{First experimental results with 50 random seeds ($\alpha$ is squashed by $\tanh(\alpha/2)$ to plot all in a unified manner)}
    \label{fig:result_toy}
\end{figure}

The obtained results from 50 random seeds are summarized in Fig.~\ref{fig:result_toy}.
Note that the horizontal axes are given as $\tanh(\alpha/2)$ to include the cases of Random and H-BoN sampling (i.e. $\alpha=\pm\infty$).
As expected, exploitation and exploration were found to be important for CartPole and PointMass tasks, respectively.
Consequently, Random sampling (with $\alpha=-\infty$) and H-BoN sampling (with $\alpha=\infty$), which are specialized for these abilities, tended to perform well, respectively.
At the same time, as suggested by EED, performance of each method was poor on the other task, as expected.
Furthermore, when $\alpha$ was varied incrementally in E-BoN sampling, a generally consistent trend was observed: performance declined/increased gradually in CartPole/PointMass task.
In other words, these results suggest that E-BoN sampling has the ability to balance EED depending on $\alpha$.
It should be noted that $\alpha=0$, which corresponds to S-BoN sampling, did not necessarily result in an optimal balance of EED; while CartPole maintained high performance, PointMass exhibited intermediate performance.
Therefore, to maximize learning performance by achieving both exploration and exploitation, simply choosing the intermediate value of $\alpha=0$ is not the best approach.

\subsection{Improvements in learning performance}

\begin{table}[ht]
    \caption{Second experimental results with eight random seeds: the values in each cell refer to the interquartile mean (interquartile range); red texts means the best results and magenta ones are the second-best.}
    \label{tab:result_locomo}
    \centering
    {\scriptsize
    \begin{tabular}{lccc}
        Strategy & Cheetah (w/ 3000 episodes) & Walker (w/ 3000 episodes) & Quadruped (w/ 6000 episodes)
        \\ \hline \\
        Random sampling & 264.47 (11.66) & 300.59 (21.49) & \textcolor{red}{438.57 (45.35)}
        \\
        H-BoN sampling & 263.61 (44.91) & \textcolor{red}{313.68 (20.98)} & 395.56 (46.83)
        \\
        S-BoN sampling & \textcolor{magenta}{264.84 (18.72)} & 290.28 (19.40) & 398.13 (106.93)
        \\
        E-BoN sampling (Proposal) & \textcolor{red}{280.97 (14.83)} & \textcolor{magenta}{301.38 (24.40)} & \textcolor{magenta}{416.37 (40.69)}
        \\
    \end{tabular}
    }
\end{table}

\begin{figure}[ht]
    \centering
    \includegraphics[keepaspectratio=true,width=0.96\linewidth]{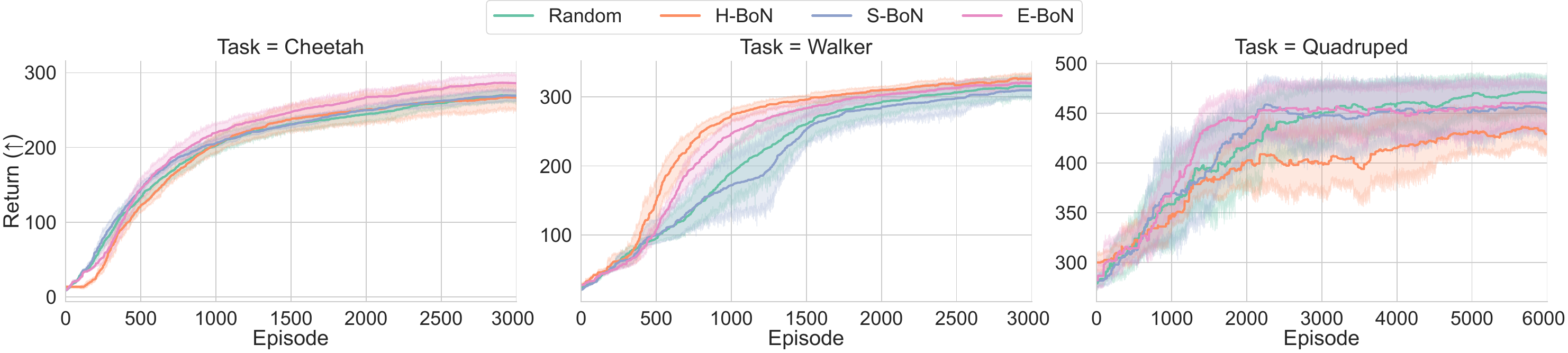}
    \caption{Learning curves for the second experiments}
    \label{fig:result_locomo}
\end{figure}

The obtained results from eight random seeds are summarized in Table~\ref{tab:result_locomo}.
In addition, their learning curves are also depicted in Fig.~\ref{fig:result_locomo}.
In Cheetah, although there is no significant difference between the baselines, the proposed E-BoN sampling successfully learned to consistently achieve superior performance.
In Walker, H-BoN sampling worked effectively, because more efficient exploration was crucial, and the proposed E-BoN sampling also learned efficiently to keep pace, securing second place.
In Quadruped, conversely, Random sampling, which prioritizes exploitation, performed best, probably due to the high number of degrees of freedom; nevertheless, the proposed E-BoN sampling was able to progress more efficiently during the early stages of learning.
On the other hand, S-BoN sampling, which is fixed at $\alpha=0$ for the intermediate EED, yielded poor results overall.
Thus, the proposed E-BoN sampling benefited from both exploration promotion through empowerment and knowledge exploitation via a policy optimized according to task-specific rewards, enabling it to demonstrate well-balanced and strong learning performance across all tasks.

\section{Conclusion}

This paper proposes a novel E-BoN sampling trick that allows for policy adjustments at reasoning rather than during training in RL.
In the conventional H-BoN sampling, computational cost fluctuates in tandem with the amount of adjustments, while S-BoN sampling is sensitive to the scale of the objective function used as a reference for adjustments.
In contrast, E-BoN sampling acquires an additional parameter through an extension based on Tsalis statistics, enabling a design where it is applied to the adjustments, with the remaining parameters are used to address the issues of each method.
Using this trick, a new framework that makes EED flexible at reasoning by introducing empowerment, which promotes exploration, as an objective function.
The experimental results indicated that the proposed E-BoN sampling acquires the expected EED flexibility and achieves well-balanced learning performance compared to the baseline methods in complex locomotion tasks.
However, since the main focus of this paper was the extension of BoN sampling, the specific EED balancing strategy remains a simple sample-based one.
In the future, a novel EED balancing strategy will be developed in a data-driven manner, especially for lifelong agents.

\subsubsection*{Acknowledgments}

This work was supported by JST PRESTO, Japan, Grant Number JPMJPR2514.

\clearpage
\bibliography{collas2026_conference}

@book{sutton2018reinforcement,
  title={Reinforcement learning: An introduction},
  author={Sutton, Richard S and Barto, Andrew G},
  year={2018},
  publisher={MIT press}
}

@article{kumar2025continual,
  title={Continual learning as computationally constrained reinforcement learning},
  author={Kumar, Saurabh and Marklund, Henrik and Rao, Ashish and Zhu, Yifan and Jeon, Hong Jun and Yueyang, Liu and Van Roy, Benjamin},
  journal={Foundations and Trends{\textregistered} in Machine Learning},
  volume={18},
  number={4},
  pages={913--1053},
  year={2025},
  publisher={Emerald Publishing Limited}
}

@article{silver2017mastering,
  title={Mastering the game of go without human knowledge},
  author={Silver, David and Schrittwieser, Julian and Simonyan, Karen and Antonoglou, Ioannis and Huang, Aja and Guez, Arthur and Hubert, Thomas and Baker, Lucas and Lai, Matthew and Bolton, Adrian and others},
  journal={nature},
  volume={550},
  number={7676},
  pages={354--359},
  year={2017},
  publisher={Nature Publishing Group UK London}
}

@inproceedings{rudin2022learning,
  title={Learning to walk in minutes using massively parallel deep reinforcement learning},
  author={Rudin, Nikita and Hoeller, David and Reist, Philipp and Hutter, Marco},
  booktitle={Conference on robot learning},
  pages={91--100},
  year={2022},
  organization={PMLR}
}

@article{hu2023simulation,
  title={How simulation helps autonomous driving: A survey of sim2real, digital twins, and parallel intelligence},
  author={Hu, Xuemin and Li, Shen and Huang, Tingyu and Tang, Bo and Huai, Rouxing and Chen, Long},
  journal={IEEE Transactions on Intelligent Vehicles},
  volume={9},
  number={1},
  pages={593--612},
  year={2023},
  publisher={IEEE}
}

@article{radosavovic2024real,
  title={Real-world humanoid locomotion with reinforcement learning},
  author={Radosavovic, Ilija and Xiao, Tete and Zhang, Bike and Darrell, Trevor and Malik, Jitendra and Sreenath, Koushil},
  journal={Science Robotics},
  volume={9},
  number={89},
  pages={eadi9579},
  year={2024},
  publisher={American Association for the Advancement of Science}
}

@article{amin2021survey,
  title={A survey of exploration methods in reinforcement learning},
  author={Amin, Susan and Gomrokchi, Maziar and Satija, Harsh and Van Hoof, Herke and Precup, Doina},
  journal={arXiv preprint arXiv:2109.00157},
  year={2021}
}

@inproceedings{klyubin2005all,
  title={All else being equal be empowered},
  author={Klyubin, Alexander S and Polani, Daniel and Nehaniv, Chrystopher L},
  booktitle={European Conference on Artificial Life},
  pages={744--753},
  year={2005},
  organization={Springer}
}

@inproceedings{bharadhwaj2022information,
  title={Information Prioritization through Empowerment in Visual Model-based RL},
  author={Bharadhwaj, Homanga and Babaeizadeh, Mohammad and Erhan, Dumitru and Levine, Sergey},
  booktitle={International Conference on Learning Representations},
  year={2022}
}

@article{schneider2025information,
  title={Information-Theoretic Policy Pre-Training with Empowerment},
  author={Schneider, Moritz and Krug, Robert and Vaskevicius, Narunas and Palmieri, Luigi and Volpp, Michael and Boedecker, Joschka},
  journal={arXiv preprint arXiv:2510.05996},
  year={2025}
}

@book{thrun1992efficient,
  title={Efficient exploration in reinforcement learning},
  author={Thrun, Sebastian B},
  year={1992},
  publisher={Carnegie Mellon University}
}

@inproceedings{plappert2018parameter,
  title={Parameter Space Noise for Exploration},
  author={Plappert, Matthias and Houthooft, Rein and Dhariwal, Prafulla and Sidor, Szymon and Chen, Richard Y and Chen, Xi and Asfour, Tamim and Abbeel, Pieter and Andrychowicz, Marcin},
  booktitle={International Conference on Learning Representations},
  year={2018}
}

@inproceedings{eberhard2023pink,
  title={Pink noise is all you need: Colored noise exploration in deep reinforcement learning},
  author={Eberhard, Onno and Hollenstein, Jakob and Pinneri, Cristina and Martius, Georg},
  booktitle={International Conference on Learning Representations},
  year={2023}
}

@article{chiappa2023latent,
  title={Latent exploration for reinforcement learning},
  author={Chiappa, Alberto Silvio and Marin Vargas, Alessandro and Huang, Ann and Mathis, Alexander},
  journal={Advances in Neural Information Processing Systems},
  volume={36},
  pages={56508--56530},
  year={2023}
}

@inproceedings{wang2023diffusion,
  title={Diffusion Policies as an Expressive Policy Class for Offline Reinforcement Learning},
  author={Wang, Zhendong and Hunt, Jonathan J and Zhou, Mingyuan},
  booktitle={International Conference on Learning Representations},
  year={2023}
}

@article{cesa2017boltzmann,
  title={Boltzmann exploration done right},
  author={Cesa-Bianchi, Nicol{\`o} and Gentile, Claudio and Lugosi, G{\'a}bor and Neu, Gergely},
  journal={Advances in neural information processing systems},
  volume={30},
  year={2017}
}

@article{chentanez2004intrinsically,
  title={Intrinsically motivated reinforcement learning},
  author={Chentanez, Nuttapong and Barto, Andrew and Singh, Satinder},
  journal={Advances in neural information processing systems},
  volume={17},
  year={2004}
}

@article{bellemare2016unifying,
  title={Unifying count-based exploration and intrinsic motivation},
  author={Bellemare, Marc and Srinivasan, Sriram and Ostrovski, Georg and Schaul, Tom and Saxton, David and Munos, Remi},
  journal={Advances in neural information processing systems},
  volume={29},
  year={2016}
}

@inproceedings{burda2019exploration,
  title={Exploration by random network distillation},
  author={Burda, Yuri and Edwards, Harrison and Storkey, Amos and Klimov, Oleg},
  booktitle={International Conference on Learning Representations},
  year={2019}
}

@inproceedings{pathak2019self,
  title={Self-supervised exploration via disagreement},
  author={Pathak, Deepak and Gandhi, Dhiraj and Gupta, Abhinav},
  booktitle={International conference on machine learning},
  pages={5062--5071},
  year={2019},
  organization={PMLR}
}

@article{yuan2022renyi,
  title={R{\'e}nyi state entropy maximization for exploration acceleration in reinforcement learning},
  author={Yuan, Mingqi and Pun, Man-On and Wang, Dong},
  journal={IEEE transactions on artificial intelligence},
  volume={4},
  number={5},
  pages={1154--1164},
  year={2022},
  publisher={IEEE}
}

@article{haarnoja2018soft2,
  title={Soft actor-critic algorithms and applications},
  author={Haarnoja, Tuomas and Zhou, Aurick and Hartikainen, Kristian and Tucker, George and Ha, Sehoon and Tan, Jie and Kumar, Vikash and Zhu, Henry and Gupta, Abhishek and Abbeel, Pieter and others},
  journal={arXiv preprint arXiv:1812.05905},
  year={2018}
}

@article{kobayashi2023reward,
  title={Reward bonuses with gain scheduling inspired by iterative deepening search},
  author={Kobayashi, Taisuke},
  journal={Results in Control and Optimization},
  volume={12},
  pages={100244},
  year={2023},
  publisher={Elsevier}
}

@inproceedings{hugessen2024surprise,
  title={Surprise-Adaptive Intrinsic Motivation for Unsupervised Reinforcement Learning},
  author={Hugessen, Adriana and Castanyer, Roger Creus and Mohamed, Faisal and Berseth, Glen},
  booktitle={Reinforcement Learning Conference},
  year={2024}
}

@inproceedings{ng1999policy,
  title={Policy Invariance Under Reward Transformations: Theory and Application to Reward Shaping},
  author={Ng, Andrew Y and Harada, Daishi and Russell, Stuart J},
  booktitle={International Conference on Machine Learning},
  pages={278--287},
  year={1999}
}

@inproceedings{morere2018bayesian,
  title={Bayesian RL for goal-only rewards},
  author={Morere, Philippe and Ramos, Fabio},
  booktitle={Conference on Robot Learning},
  pages={386--398},
  year={2018},
  organization={PMLR}
}

@inproceedings{al2022getting,
  title={Getting priorities right: Intrinsic motivation with multi-objective reinforcement learning},
  author={Al-Husaini, Yusuf and Rolf, Matthias},
  booktitle={IEEE International Conference on Development and Learning},
  pages={208--214},
  year={2022},
  organization={IEEE}
}

@article{stiennon2020learning,
  title={Learning to summarize with human feedback},
  author={Stiennon, Nisan and Ouyang, Long and Wu, Jeffrey and Ziegler, Daniel and Lowe, Ryan and Voss, Chelsea and Radford, Alec and Amodei, Dario and Christiano, Paul F},
  journal={Advances in neural information processing systems},
  volume={33},
  pages={3008--3021},
  year={2020}
}

@inproceedings{verdun2025soft,
  title={Soft Best-of-$ n $ Sampling for Model Alignment},
  author={Verdun, Claudio Mayrink and Oesterling, Alex and Lakkaraju, Himabindu and Calmon, Flavio P},
  booktitle={IEEE International Symposium on Information Theory},
  pages={1--6},
  year={2025},
  organization={IEEE}
}

@article{ichihara2025evaluation,
  title={Evaluation of Best-of-N Sampling Strategies for Language Model Alignment},
  author={Ichihara, Yuki and Jinnai, Yuu and Morimura, Tetsuro and Abe, Kenshi and Ariu, Kaito and Sakamoto, Mitsuki and Uchibe, Eiji},
  journal={Transactions on Machine Learning Research},
  year={2025}
}

@inproceedings{peters2019sparse,
  title={Sparse sequence-to-sequence models},
  author={Peters, Ben and Niculae, Vlad and Martins, Andr{\'e} FT},
  booktitle={Annual Meeting of the Association for Computational Linguistics},
  pages={1504--1519},
  year={2019}
}

@inproceedings{gonccalves2025adasplash,
  title={AdaSplash: Adaptive Sparse Flash Attention},
  author={Gon{\c{c}}alves, Nuno and Treviso, Marcos V and Martins, Andre},
  booktitle={International Conference on Machine Learning},
  pages={19878--19896},
  year={2025},
  organization={PMLR}
}

@article{tsallis1988possible,
  title={Possible generalization of Boltzmann-Gibbs statistics},
  author={Tsallis, Constantino},
  journal={Journal of statistical physics},
  volume={52},
  number={1-2},
  pages={479--487},
  year={1988},
  publisher={Springer}
}

@article{tunyasuvunakool2020dm_control,
  title={dm\_control: Software and tasks for continuous control},
  author={Tunyasuvunakool, Saran and Muldal, Alistair and Doron, Yotam and Liu, Siqi and Bohez, Steven and Merel, Josh and Erez, Tom and Lillicrap, Timothy and Heess, Nicolas and Tassa, Yuval},
  journal={Software Impacts},
  volume={6},
  pages={100022},
  year={2020},
  publisher={Elsevier}
}

@software{jun2023shimmy,
  author={Jun Jet Tai and Mark Towers and Elliot Tower},
  title={{Shimmy: Gymnasium and PettingZoo Wrappers for Commonly Used Environments}},
  year={2023},
  publisher={Zenodo},
  version={v1.1.0},
  doi={10.5281/zenodo.8140744},
}

@inproceedings{haarnoja2018soft,
  title={Soft actor-critic: Off-policy maximum entropy deep reinforcement learning with a stochastic actor},
  author={Haarnoja, Tuomas and Zhou, Aurick and Abbeel, Pieter and Levine, Sergey},
  booktitle={International conference on machine learning},
  pages={1861--1870},
  year={2018},
  organization={Pmlr}
}

@article{nielsen2020non,
  title={Non-negative Monte Carlo estimation of f-divergences},
  author={Nielsen, Frank},
  year={2020}
}

@inproceedings{akiba2019optuna,
  title={Optuna: A next-generation hyperparameter optimization framework},
  author={Akiba, Takuya and Sano, Shotaro and Yanase, Toshihiko and Ohta, Takeru and Koyama, Masanori},
  booktitle={ACM SIGKDD international conference on knowledge discovery \& data mining},
  pages={2623--2631},
  year={2019}
}

@article{peel2000robust,
  title={Robust mixture modelling using the t distribution},
  author={Peel, David and McLachlan, Geoffrey J},
  journal={Statistics and computing},
  volume={10},
  number={4},
  pages={339--348},
  year={2000},
  publisher={Springer}
}

@article{ridders2003new,
  title={A new algorithm for computing a single root of a real continuous function},
  author={Ridders, C},
  journal={IEEE Transactions on circuits and systems},
  volume={26},
  number={11},
  pages={979--980},
  year={2003},
  publisher={IEEE}
}

@article{zhang2019root,
  title={Root mean square layer normalization},
  author={Zhang, Biao and Sennrich, Rico},
  journal={Advances in Neural Information Processing Systems},
  volume={32},
  pages={12381--12392},
  year={2019}
}

@article{barron2021squareplus,
  title={Squareplus: A Softplus-Like Algebraic Rectifier},
  author={Barron, Jonathan T},
  journal={arXiv preprint arXiv:2112.11687},
  year={2021}
}

@inproceedings{kobayashi2024consolidated,
  title={Consolidated adaptive t-soft update for deep reinforcement learning},
  author={Kobayashi, Taisuke},
  booktitle={International Joint Conference on Neural Networks},
  pages={1--8},
  year={2024},
  organization={IEEE}
}

@article{ilboudo2023adaterm,
  title={Adaterm: Adaptive t-distribution estimated robust moments for noise-robust stochastic gradient optimization},
  author={Ilboudo, Wendyam Eric Lionel and Kobayashi, Taisuke and Matsubara, Takamitsu},
  journal={Neurocomputing},
  volume={557},
  pages={126692},
  year={2023},
  publisher={Elsevier}
}

@article{han2021max,
  title={A max-min entropy framework for reinforcement learning},
  author={Han, Seungyul and Sung, Youngchul},
  journal={Advances in Neural Information Processing Systems},
  volume={34},
  pages={25732--25745},
  year={2021}
}

@inproceedings{nauman2025decoupled,
  title={Decoupled policy actor-critic: Bridging pessimism and risk awareness in reinforcement learning},
  author={Nauman, Michal and Cygan, Marek},
  booktitle={AAAI Conference on Artificial Intelligence},
  volume={39},
  number={18},
  pages={19633--19641},
  year={2025}
}
\bibliographystyle{collas2026_conference}

\appendix
\section{State transition models}
\label{app:model}

To implement empowerment, it is necessary to model the state transition probabilities with and without the condition by action, $p_e$ and $p_m$, respectively.
This paper does not prioritize model accuracy; rather, models that can rapidly compute the exact log-likelihood and the analytical solution for the distribution mean (or sampling).
Under this consideration, $p_e$ is modeled by a single Student’s t-distribution.
In addition, $p_m$ must encompass multiple distributions arising from differences in action, it is modeled by a mixture of t-distributions with 10 components.
All model parameters are output from the respective MLPs that take the state (and action) as input, which are the same design as the policy and value functions.
However, to ensure that the mean can be obtained in closed form, the degrees of freedom parameter, one of the parameters of the t-distribution, is set to 2 or greater (to ensure that the variance can also be obtained, just to be safe).
Note that a Gaussian distribution was not used because the t-distribution can more robustly represent multimodality by treating data from different components as outliers \citep{peel2000robust}.
These models are trained using data sampled by experience replay in RL, with the negative log-likelihood serving as the loss function.

\section{Approximated solution of entmax}
\label{app:entmax}

The probability $P_i$ at $entmax$ in eq.~\eqref{eq:res_entmax} requires that $\lambda$ be specified.
Since the q-exponential function is monotonically increasing, $P_i$ for $i=1,\ldots,N$ decreases monotonically w.r.t. $\lambda$.
Furthermore, since $\sum_{i=1}^N P_i = 1$ holds, $\lambda$ can generally be obtained numerically using a root-finding algorithm.
In the original work, the bisection algorithm was adopted because it is simple, numerically stable, and converges asymptotically.
In later improvements, a method combining the bisection algorithm with the Halley algorithm was proposed; however, both are iterative solvers, and their computational cost is not constant.
Therefore, this paper develops an alternative method to approximate $\lambda$ with a constant computational cost.

Specifically, the upper and lower bounds for $\lambda$, $[\underline{\lambda}, \overline{\lambda}]$, are first derived.
With $\beta J(x_i) = J_i$, in the conventional approach, these are given using the following inequality constraints:
\begin{align}
    & \exp_{1-\alpha}(\overline{J} - \lambda) \leq \sum_{i=1}^N P_i = 1 \leq N \exp_{1-\alpha}(\overline{J} - \lambda)
    \nonumber \\
    \Rightarrow & \overline{J} \leq \lambda \leq \overline{J} - \ln_{1-\alpha} \frac{1}{N} = \overline{J} + \ln_{\alpha+1} N
\end{align}
where, $\overline{J} = \max_i J_i$.
Note that the original paper does not explicitly deal with the q-exponential function, and since the definition of $\lambda$ differs slightly from that used in this paper, the result above may appear to have slightly different bounds; however, upon transformation, they are equivalent.

In addition, to make the bounds tighter, let's reconsider $\underline{\lambda}$.
First, using the fact that $\exp_{q_1}(x) \geq \exp_{q_2}(x)$ for $q_1 \geq q_2$, another lower bound that holds only when $\alpha \leq 0$ is given as follows:
\begin{align}
    \sum_{i=1}^N \exp(J_i - \lambda) \leq \sum_{i=1}^N P_i = 1
    \Rightarrow \mathrm{LSE}(J_{1:N}) \leq \lambda \quad (\alpha \leq 0)
\end{align}
where, $\mathrm{LSE}(J_{1:N}) = \ln \sum_{i=1}^N \exp(J_i)$ denotes the log-sum-exponential function, which is larger than or equivalent to the maximum function.
Next, using $\underline{J} = \min_i J_i$ yields the following condition.
\begin{align}
    N \exp_{1-\alpha}(\underline{J} - \lambda) \leq \sum_{i=1}^N P_i = 1
    \Rightarrow \underline{J} + \ln_{\alpha+1} N \leq \lambda
\end{align}
Whether this lower bound is greater than $\overline{J}$ and $\mathrm{LSE}(J_{1:N})$ depends on $J_i$, so a conditional branch is necessary.
In summary, $\underline{\lambda}$ is more tightly derived in this paper.
\begin{align}
    \underline{\lambda} = \begin{cases}
        \max \{ \mathrm{LSE}(J_{1:N}), \underline{J} + \ln_{\alpha+1} N \} & \alpha \leq 0
        \\
        \max \{ \overline{J}, \underline{J} + \ln_{\alpha+1} N \} & \alpha > 0
    \end{cases}
\end{align}

Similarly, another upper bound for $\alpha \geq 0$ is given using $\mathrm{LSE}(J_{1:N})$.
\begin{align}
    \sum_{i=1}^N P_i = 1 \leq \sum_{i=1}^N \exp(J_i - \lambda)
    \Rightarrow \lambda \leq \mathrm{LSE}(J_{1:N}) \quad (\alpha \geq 0)
\end{align}
Unfortunately, since $\ln_{\geq 1}(x) \leq \ln(x)$, it follows that $\overline{J} + \ln_{\alpha+1} N \leq \overline{J} + \ln N$ holds for $\overline{J} + \ln N$, which is an upper bound of $\mathrm{LSE}(J_{1:N})$.
That is, $\overline{\lambda}$ is determined by the following conditional branching.
\begin{align}
    \overline{\lambda} = \begin{cases}
        \overline{J} + \ln_{\alpha+1} N & \alpha < 0
        \\
        \min \{ \overline{J} + \ln_{\alpha+1} N, \mathrm{LSE}(J_{1:N}) \} & \alpha \geq 0
    \end{cases}
\end{align}
Note that when $\alpha = 0$, the range of $\lambda$ is restricted to a single point, $\lambda = \mathrm{LSE}(J_{1:N})$, which correctly satisfies the softmax function.
This cannot be achieved by the conventional bounds due to ignoring the property of q-exponential function, which enables to find the inequalities for $\mathrm{LSE}(J_{1:N})$.

From the obtained $[\underline{\lambda}, \overline{\lambda}]$, $\lambda$ is appropriately approximated without performing iterative calculations.
Specifically, let's define the error function $e(\lambda) = \ln \sum_{i=1}^N \exp_{1-\alpha}(J_i - \lambda)$.
If this equals $1$, $\lambda$ is optimal; furthermore, given the monotonicity of the function, $e(\underline{\lambda}) \geq 0$ and $e(\overline{\lambda}) \leq 0$ hold.
Therefore, $e_0 = e(\lambda_0)$ for $\lambda_0 = (\underline{\lambda} + \overline{\lambda}) / 2$ is first computed to check its sign.
The pair of it, $\lambda_2$, is selected as $\overline{\lambda}$ if the sign is positive or $\underline{\lambda}$ if negative.
The corresponding $e_2 = e(\lambda_2)$ is also computed.
The midpoint $\lambda_1 = (\lambda_0 + \lambda_2) / 2$ is finally given with its error $e_1 = e(\lambda_1)$.
With these preparation, $\lambda$ is approximated with the next-point estimation by Ridders method \citep{ridders2003new}.
\begin{align}
    \lambda &\simeq \lambda_1 + (\lambda_1 - \lambda_0) \frac{\mathrm{sign}(e_0) e_1}{\sqrt{e_1^2 - e_0 e_2}}
    \nonumber \\
    &= \frac{1}{4} \left[ \{C + (2 + \mathrm{sign}(e_0))\} (\overline{\lambda} - \underline{\lambda}) - 4\underline{\lambda}\right]
\end{align}
where, $C = e_1 / \sqrt{e_1^2 - e_0 e_2}$.
Although the computation of the error function accounts for the majority of the computational cost, this method completes it in three times, achieving a computational cost comparable to that of the conventional bisection method after approximately three iterations.
Furthermore, by leveraging monotonic decreasing properties to halve the search range without computing the error function, and then applying the highly convergent Ridders method, we can expect sufficient approximation performance.

The effectiveness of the above implementation is verified through two experiments, where random variables $x_i$ ($i=1,\ldots,N$) are sampled from Gaussian with scale $\sigma$ and the absolute error, $|e|$, and the computational time are evaluated.

To evaluate statistical performance under various conditions, 1000 combinations are given: $N=\{4, 16, 64, 256, 1024\}$, $\sigma=\{0.01, 0.1, 1, 10, 100\}$, and $\alpha = [-2, 2]$ with $0.1$ increments (except $\alpha = 0$ since it is softmax).
For each of the 1000 combinations, $100$-sample batch computation is conducted to quantify the results using the interquartile mean of $100$ samples.
First experiment is to verify the tighter range of $\lambda$ with the conventional bisection method.
Using the tighter range, second experiment changes the solvers to show the quality and efficiency of approximation.

\begin{figure}[ht]
    \begin{subfigure}[b]{0.48\linewidth}
        \centering
        \includegraphics[keepaspectratio=true,width=\linewidth]{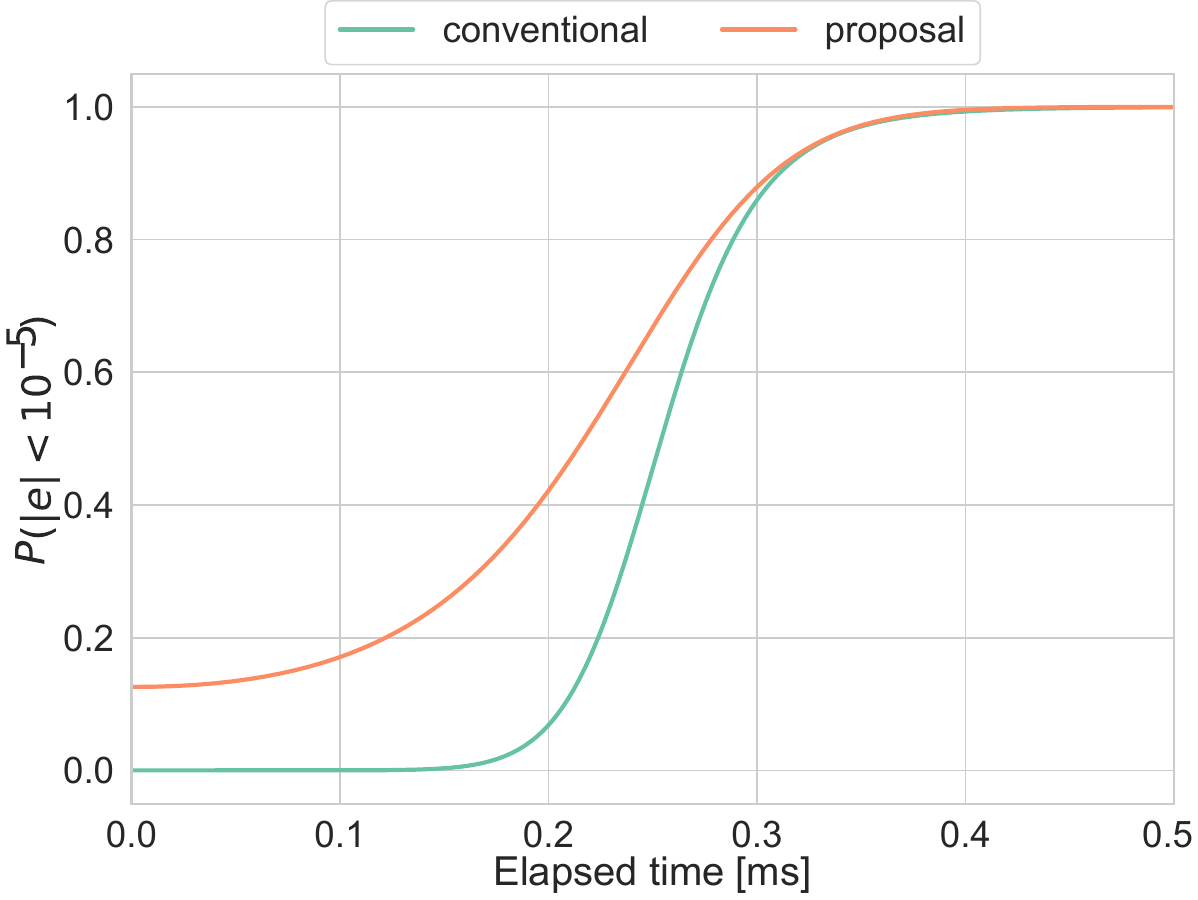}
        \subcaption{Comparison of bounds}
        \label{fig:entmax_bound}
    \end{subfigure}
    \begin{subfigure}[b]{0.48\linewidth}
        \centering
        \includegraphics[keepaspectratio=true,width=\linewidth]{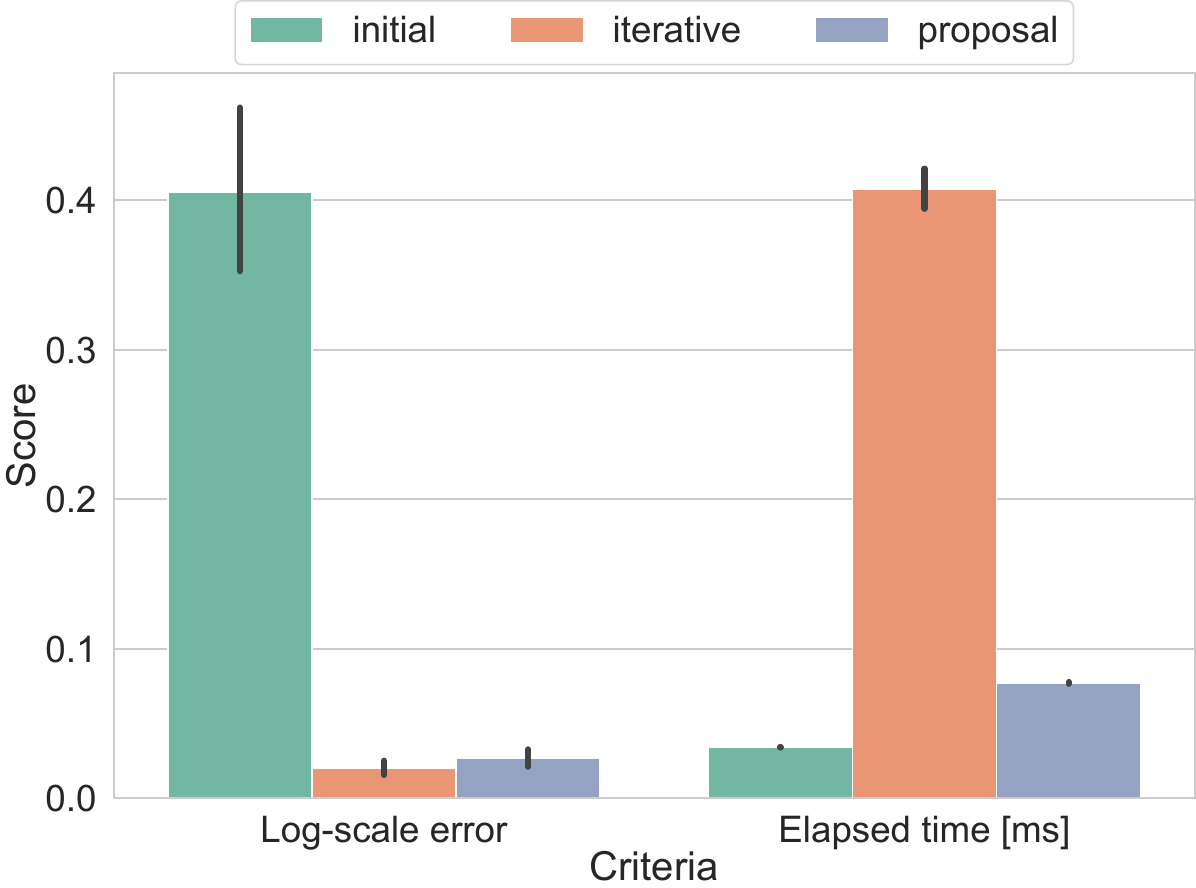}
        \subcaption{Comparison of solvers}
        \label{fig:entmax_solver}
    \end{subfigure}
    \caption{Evaluation of the proposed approximation for entmax}
    \label{fig:entmax}
\end{figure}

The experimental results are summarized in Fig.~\ref{fig:entmax}.
Note that these experiments were conducted on a MacBook Pro equipped with an Apple M4 CPU and 24~GB of RAM.
The left chart corresponds to the results of the first experiment and depicts the probability that the absolute error w.r.t. the cumulative computation time, as the number of iterations in the bisection method increases (capped at 30), falls below the small threshold ($10^{-5}$ in this implementation).
As can be seen at a glance, while convergence could never be confirmed in the early stages of iteration within the conventional range, the derived tighter one enables convergence in just a few iterations.
Although the tighter range requires additional computational costs (e.g. for computing log-sum-exponential function), this result sufficiently justifies them.
On the other hand, the right chart shows the results of the second experiment, confirming that the proposed approximation achieves an error level comparable to that of the conventional method by limiting the number of iterations with convergence checks.
In this case, there is a more than fourfold difference in computation time (though this includes the overhead from convergence checks required to achieve comparable error levels).
Furthermore, while the conventional method exhibits a wide confidence interval for computation time due to fluctuations in the number of iterations, the proposed method maintains a very narrow confidence interval due to its constant computational cost.
Compared to the initial estimate $\lambda = (\underline{\lambda} + \overline{\lambda})/2$, the proposed method limits the error to 10~\% or less while keeping the increase in computation time to approximately twofold, resulting in a sufficiently fast execution of less than 0.1~ms.
Based on the above, we conclude that the proposed entmax approximation method successfully balances sufficient accuracy with a constant computational cost.
Note that in the actual implementation, $P_i$ needs to be divided by its summation to forcibly satisfy the definition of probability and ccompensate for the remaining error.

\section{Hyperparameters of implementation}
\label{app:config}

As mentioned in Section~\ref{subsec:sac}, the baseline RL algorithm in this paper is SAC.
It has three network models: one is for a policy; and the remaining two are for action-value functions.
All the models have MLP with two layers and 100 neurons for each.
Their activation function is RMSNorm \citep{zhang2019root} and squish function \citep{barron2021squareplus}, which is an alternative of swish function.
As for the policy's distribution model, since the basic squashing with tanh function loses the analytical mean, this paper employs modified PERT distribution with its mode and sharpness parameters.
The greedy output for tests is given by its analytical mean.
As for the action-value function, two separated models are independently trained.
They have the corresponding target networks, which are updated by CAT-soft update \citep{kobayashi2024consolidated}.

The target value is given by the minimum value of two target networks.
Note that to stabilize learning, max-min entropy (MME) \citep{han2021max} is also employed: namely, the positive log-likelihood of policy for the next state is added to the target value, although the basic implementation uses the negative one.
Its gain is designed as $\alpha (1 - \gamma)$.
In addition, according to the report that the policy update of SAC is too conservative due to the use of gradient for the minimum action-value function \citep{nauman2025decoupled}, this paper simply uses the maximum action-value function for computing policy gradient.
For the temperature parameter $\alpha^\mathrm{SAC}$, the automatic tuning method \citep{haarnoja2018soft2} is implemented, setting the target lower bound to $|\mathcal{A}|\ln (0.2)$ relative to the upper bound of the policy entropy, $|\mathcal{A}|\ln 2$.

These network models (and the temperature parameter) are optimized using AdaTerm \citep{ilboudo2023adaterm}, a variant of stochastic gradient descent that is robust to gradient noise and outliers.
Its learning rate is set to $10^{-4}$, with all other settings left at their defaults.
Training is scheduled at the end of each episode, at which point half of the experience data in the replay buffer (capacity $102400$) is divided into batches (size $256$) and randomly replayed.
Additionally, $N$ in BoN sampling is set to $256$.

\end{document}